\documentclass[times, review, 10pt]{elsarticle}

\usepackage{amssymb}
\usepackage{amsmath}
\usepackage{subfiles}
\usepackage{booktabs}
\usepackage{multirow}
\usepackage{rotating}
\usepackage{adjustbox}
\usepackage{pifont} 
\usepackage{soul}
\usepackage[table,dvipsnames]{xcolor}
\usepackage{color}
\colorlet{tdcolor}{yellow!35}
\sethlcolor{tdcolor}

\journal{Pattern Recognition}

\begin{document}

\begin{frontmatter}



\title{Text-guided Weakly Supervised Framework for \\ Dynamic Facial Expression Recognition}


\author[label1]{Gunho Jung}
\ead{gh\_jung@korea.ac.kr}

\author[label2]{Heejo Kong}
\ead{hj\_kong@korea.ac.kr}
         
\author[label1]{Seong-Whan Lee\corref{cor1}}
\ead{sw.lee@korea.ac.kr}

\affiliation[label1]{organization={Department of Artificial Intelligence, Korea University},
            city={Seoul},
            postcode={02841}, 
            country={Republic of Korea}}
            
\affiliation[label2]{organization={Department of Brain and Cognitive Engineering, Korea University},
            city={Seoul},
            postcode={02841}, 
            country={Republic of Korea}}
\cortext[cor1]{Corresponding author.}
\begin{abstract}
Dynamic facial expression recognition (DFER) aims to identify emotional states by modeling the temporal changes in facial movements across video sequences. A key challenge in DFER is the many-to-one labeling problem, where a video composed of numerous frames is assigned a single emotion label. A common strategy to mitigate this issue is to formulate DFER as a Multiple Instance Learning (MIL) problem. However, MIL-based approaches inherently suffer from the visual diversity of emotional expressions and the complexity of temporal dynamics. To address this challenge, we propose TG-DFER, a text-guided weakly supervised framework that enhances MIL-based DFER by incorporating semantic guidance and coherent temporal modeling. We incorporate a vision-language pre-trained (VLP) model is integrated to provide semantic guidance through fine-grained textual descriptions of emotional context. Furthermore, we introduce visual prompts, which align enriched textual emotion labels with visual instance features, enabling fine-grained reasoning and frame-level relevance estimation. In addition, a multi-grained temporal network is designed to jointly capture short-term facial dynamics and long-range emotional flow, ensuring coherent affective understanding across time. Extensive results demonstrate that TG-DFER achieves improved generalization, interpretability, and temporal sensitivity under weak supervision.

\end{abstract}




\end{frontmatter}


\section{Introduction}

Facial expressions serve as direct indicators of human emotions, playing a crucial role in interpreting feelings during human interactions \cite{bib6, bib8}. Recognizing these expressions is essential in various fields, including human–computer interaction (HCI) \cite{bib2}, health assessment \cite{bib3}, and driver assistance systems \cite{bib4}. Dynamic facial expression recognition (DFER) involves classifying emotions inferred from video clips. \cite{bib9, bib11}. Unlike static facial expression recognition (SFER) which focuses on individual images \cite{bib6, bib8}, DFER captures temporal facial variations across consecutive video frames, allowing a more comprehensive understanding of subtle and transient emotional cues. Consequently, DFER has emerged as a prominent research area, significantly driven by large-scale supervised learning methodologies utilizing in-the-wild video datasets. The recent availability of large annotated datasets such as DFEW \cite{bib59} and FERV39K \cite{bib60} has facilitated training robust models capable of generalizing across diverse emotional expressions.

While large annotated datasets have been instrumental in advancing DFER by supporting training on diverse and realistic emotional expressions \cite{bib21, bib23, bib25, bib26}, most provide only clip-level annotations that assign a single emotion label to an entire video segments, resulting in a many-to-one labeling problem. Since frame-level annotation is prohibitively expensive, the availability of densely labeled data remains limited. One effective approach to mitigating this limitation is to reformulate DFER as a multiple instance learning (MIL) \cite{bib75, bib76, bib78}, in which each video is treated as a bag of frame-level instances. Under this paradigm, the model learns to focus on instances that effectively convey the target emotion, suppressing the impact of irrelevant or ambiguous frames through weak supervisory signals.

\begin{figure}[t]
\centering
\includegraphics[width=0.95\linewidth]{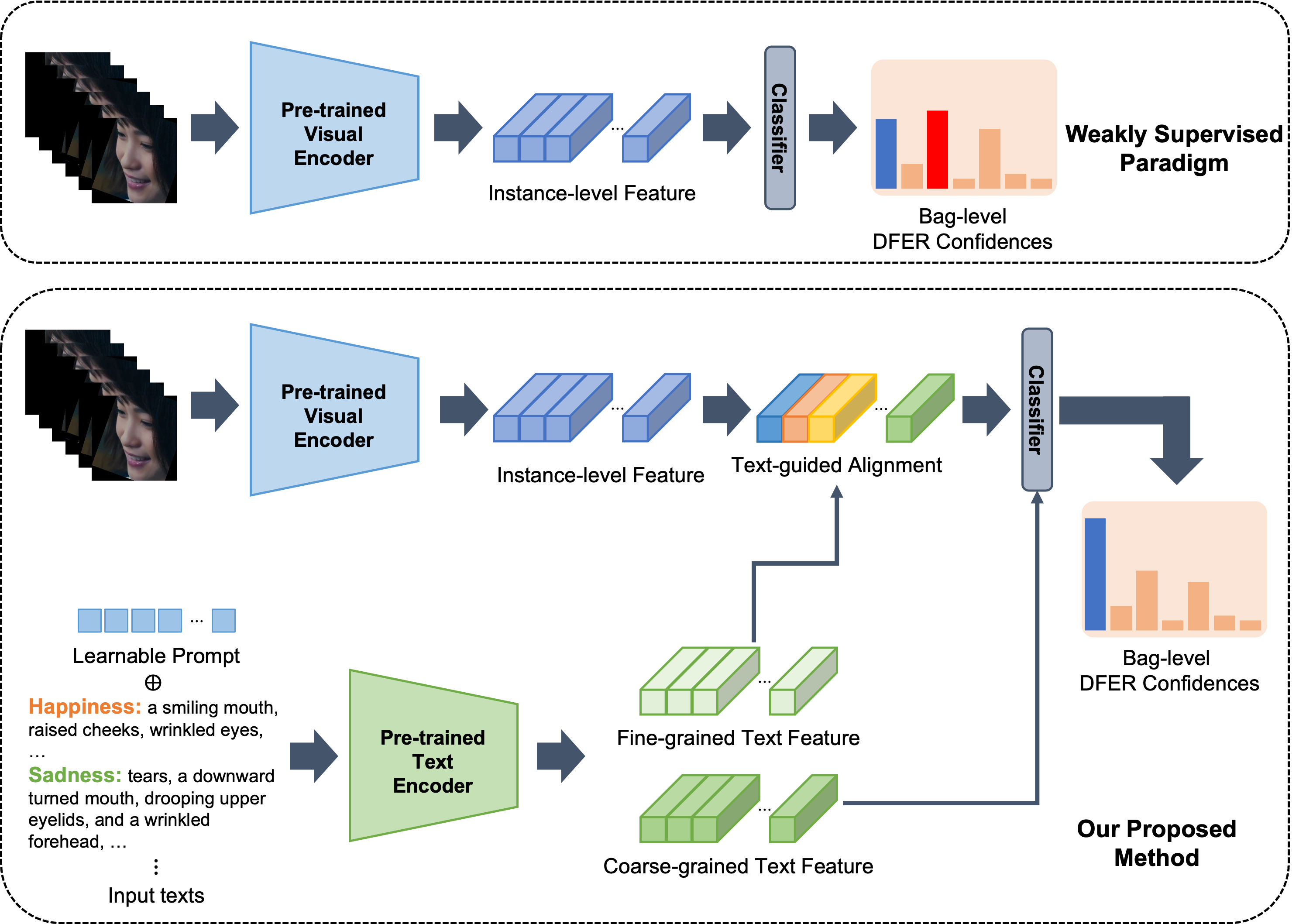}
\caption{Overview of the proposed text-guided weakly supervised framework for DFER. The weakly supervised paradigm operates under coarse clip-level supervision. In contrast, our method incorporates text-guided alignment by injecting semantic priors via pre-trained visual and text encoders.}
\label{fig_1}
\end{figure}

Despite their potential, MIL-based approaches face inherent limitations in DFER due to the visual diversity and temporal dynamics characteristic of emotional expressions. 
The high intra-class variability characteristic of emotional expressions results in widely scattered relevant instances within the feature space, thereby hindering precise discrimination between emotionally salient frames and visually comparable but emotionally irrelevant ones. 
This issue is further compounded by ambiguous frames that lack definitive emotional content, consequently reducing the reliability of predictions based on coarse-level supervision. 
In addition, discrete temporal segmentation, which divides the video into fixed segments processed independently, disrupts the continuity of emotional expression. 
Processing video segments independently limits the model's ability to capture gradual affective transitions or context-dependent cues.
Ultimately, this approach fragments the emotional flow and hinders a coherent understanding over time.


To address these limitations, we propose TG-DFER, a text-guided weakly supervised framework for dynamic facial expression recognition. 
While prior works utilize fine-grained text as a static guide, our framework introduces a novel visual prompt mechanism that creates a bi-directional feedback process, dynamically refining the textual guidance based on the video's specific visual content.

The proposed method is grounded in a weakly supervised MIL paradigm in which each video clip is treated as a bag of frame-level instances and supervised using coarse clip-level labels. To effectively capture the diverse and subtle characteristics of emotional expressions, we integrate a vision-language pre-trained (VLP) model that provides semantic guidance through textual descriptions explicitly encoding fine-grained emotional context. Furthermore, to enhance cross-modal alignment and discriminative capacity, we introduce visual prompts, which are textual emotion labels enriched with visual context and aligned with instance-level visual features. This alignment enables the model to assess the relative contribution of each frame to the overall emotional state, thereby improving both temporal sensitivity and interpretability. Instead of relying solely on text as a supplementary signal, we construct enhanced label features that strengthen the complementarity between visual and textual information, ultimately improving the model’s generalization under weak supervision.

In addition, this study introduces a multi-grained temporal network designed to capture the temporal continuity and gradual transitions of affective expressions in video sequences. The proposed architecture leverages multiple temporal granularities to simultaneously capture fine-grained facial changes over short durations and the broader affective progression across the entire video. By jointly modeling local temporal dynamics and global contextual flow, the network enables the recognition of both momentary emotional cues and their evolution within a coherent temporal structure. As a result, the model is able to identify meaningful emotional patterns not as isolated snapshots, but as part of a continuous emotional stream, thereby facilitating more accurate and reliable affective understanding.

The main contributions of this study are summarized as follows:
\begin{itemize}
\item{We propose TG-DFER, a weakly supervised framework that models DFER as a multiple instance learning problem using coarse video-level labels. By integrating a vision-language pre-trained model and aligning visual prompts with instance-level features, the framework enables fine-grained emotion localization and enhances interpretability.}

\item{We design a multi-grained temporal network that captures both short-term facial changes and long-term emotional transitions, allowing the model to preserve temporal coherence and recognize dynamic affective patterns more effectively.}

\item{We demonstrate the effectiveness of our method through extensive ablation studies and visualization results. Our approach significantly outperforms the baseline model, achieving state-of-the-art results on two widely used DFER benchmarks.}

\end{itemize}

\section{Related Works}
\subsection{Dynamic Facial Expression Recognition}

In the evolution of deep learning-based methodologies, there has been a notable improvement in FER performance over traditional approaches that leverage hand-crafted features \cite{bib70}, \cite{bib71}. Unlike SFER, which processes a singular image, DFER requires an understanding of the inter-frame relationships within a video sequence. Consequently, recent research has predominantly focused on the objective of learning both the temporal and robust spatial features of facial expressions. To achieve this, several studies \cite{bib11}, \cite{bib14} adopted Convolutional Neural Network (CNN)-based models such as ResNet and VGG to extract spatial features from individual frames. Subsequently, Recurrent Neural Network (RNN)-based models, including Long Short-Term Memory (LSTM) networks and Gated Recurrent Units (GRU), have been utilized to decipher the temporal relationships among these frames. In addition, the introduction of 3D Convolutional Neural Networks (3DCNNs) offers an innovative approach, providing a framework to learn the spatial and temporal features in a more integrated manner \cite{bib18}, \cite{bib19}.

In recent years, researchers have adopted transformers \cite{bib5} to model the long-range relationships between frames. Specifically, Zhao \emph{et al.} \cite{bib9} introduced the Dynamic Facial Expression Recognition Transformer (Former-DFER) that integrates a Convolutional Transformer (CS-Former) for spatial information learning with a Temporal Transformer (T-Former) for temporal information acquisition. Ma \emph{et al.} \cite{bib21} developed a Spatial-Temporal Transformer to learn the distinctive features of individual frames and the inter-frame relationships concurrently. Liu \emph{et al.} \cite{bib22} presented the Expression Snippet Transformer (EST) as a solution to the challenge posed by the minimal facial expression movements in videos, which are often too slight to convey meaningful spatio-temporal relationships. 
Since then, various methodologies have emerged, including noise robust learning \cite{bib23}, intensity-aware loss \cite{bib25}, weakly-supervised learning \cite{bib24}, and advanced self-supervised methods \cite{bib26} like heatmap neighbor contrastive learning \cite{bib86}. 
Other sophisticated approaches have focused on capturing long-range geometric patterns \cite{bib87} through generative models and disentangling features for domain generalization \cite{bib88}.

Recent advancements have enabled the adaptation of transformer-based methodologies, originally developed for Natural Language Processing (NLP), to computer vision tasks, thereby enhancing the extraction of both spatial and temporal information. By leveraging the substantial benefits of contrastive language-image pre-training methods, such as CLIP \cite{bib29}, for visual representation learning, the transfer of knowledge from CLIP to DFER has set new state-of-the-art performance benchmarks. Li \emph{et al.} \cite{bib27} introduced CLIPER, a model that harnesses a visual-language framework to achieve leading performance in the facial expression recognition domain. In parallel, Zhao \emph{et al.} \cite{bib28} proposed a temporal model that captures subtle facial expressions in videos through detailed descriptors of specific movements and muscle activations. Despite these advancements, ambiguity issues persist due to the variability in expression intensity within videos. Differentiating between target and non-target frames for ambiguous expressions without positional labels remains highly challenging. Furthermore, existing approaches employ simple temporal networks that are insufficient for capturing the complex dynamics of evolving expressions. In contrast, the proposed TG-DFER addresses these issues by suppressing the influence of non-relevant frames while enhancing that of highly relevant ones, and by processing video sequences at multiple temporal resolutions.

\subsection{Vision-Language Pre-Training with Text Prompting}

Vision-language models have seen remarkable progress in enhancing zero-shot performance through contrastive learning with image-text pairs, exploiting the abundant and virtually limitless resources available on the internet for a highly efficient and cost-effective strategy. The most representative vision-language models, namely the CLIP and ALIGN \cite{bib30} frameworks, have significantly advanced the state-of-the-art in a various computer vision tasks \cite{bib32}, \cite{bib33} by utilizing massive datasets consisting of 400 million and 1.8 billion image-text pairs, respectively. Subsequently, there have been attempts to extend vision-language pre-training models to the video domain. CLIP4Clip \cite{bib48} adapts the CLIP model for video-to-text retrieval tasks to handle sequences of video frames and effectively compute the similarity between video and text representations. Liu et al. \cite{bib49} proposed a Spatial-Temporal Auxiliary Network (STAN) focusing on both high-level semantic and low-level visual patterns to improve the performance of video-text retrieval and video recognition. Open-VCLIP \cite{bib50} extends CLIP to video understanding by integrating temporal attention mechanisms to improve the ability of the model to aggregate temporal features, making it suitable for video classification and zero-shot video tasks. Ma et al. \cite{bib51} introduced X-CLIP to enhance video-text retrieval by considering various levels of granularity and using attention mechanisms to filter out irrelevant information. 

Fine-tuning has been widely used to leverage large pre-trained models for downstream tasks. However, this approach requires significant computational resources owing to the need to update all model parameters. Recently, prompt tuning \cite{bib52}, \cite{bib56} was introduced into NLP as an efficient method for applying large pre-trained language models to downstream tasks. Context Optimization (CoOp) \cite{bib54} replaces hand-crafted prompts (e.g., “a photo of a [CLASS]”) with learnable soft prompts derived from a small set of labeled examples, addressing the limitations of manually created prompts that do not consider specific task knowledge. Conditional Context Optimization (CoCoOp) \cite{bib55} extends CoOp by optimizing textual prompts under various conditions and not simply the input image. 
Recent works based on CLIP in FER use descriptions of learnable contexts and expressions. 
DFER-CLIP \cite{bib28} uses LLM-generated text to create a static semantic target for video features. 
Similarly, MPA-FER \cite{bib85} employs these descriptions as a static guide to train prompts for static images. 
However, their reliance on this text as static source of knowledge presents a key limitation.
This fixed guidance fails to account for the wide visual diversity of expressions within the same emotion category. 
To overcome this issue, our TG-DFER framework introduces a novel dynamic approach, visual prompt module, ensuring the semantic guidance adapted to the specific visual context of each video.

\subsection{Multiple Instance Learning}
Multiple Instance Learning (MIL) is a learning approach where labels are assigned at the bag level rather than the instance level \cite{bib75}. A bag is labeled as negative only if all its instances are negative, and it is considered positive if at least one instance is positive. This property has proven effective in addressing label uncertainty for weakly supervised object detection (WSOD) \cite{bib76}, anomaly detection \cite{bib68}, and whole-slide image classification \cite{bib78}. In the emotion recognition domain, Romeo et al. \cite{bib82} utilized MIL for emotion detection from physiological signals, while Chen et al. \cite{bib83} employed it for pain detection, both in controlled environments or with handcrafted features. However, these approaches do not explicitly address DFER in more complex, real-world settings.

In contrast, in-the-wild DFER must contend with multidimensional noise factors, including rapidly changing emotional expressions, complex backgrounds, and variations in lighting and pose. The first approach \cite{bib24} to employ MIL for DFER modeled short-term temporal relationships during feature extraction and incorporated long-term relationships during instance fusion, thereby managing unbalanced temporal dependencies within the MIL pipeline. Building on this paradigm, our method mitigates noise and ambiguity commonly observed in MIL by explicitly clarifying the context of facial expressions through frame-level text prompts, while simultaneously capturing the overall emotional flow of the video via a multi-grained temporal network.

\section{Methodology}

\subsection{Problem Definition}
The goal of DFER under a weakly supervised paradigm is to train a classification model capable of predicting frame-level emotion confidences, given only video-level emotion labels. In other words, each video is assigned a single coarse emotion label, and no frame-level annotations are provided. Given a video \(V\) and its assigned coarse label \(y\) (e.g., "happy"), all frames in \(V\) are treated as expressing the emotion \(y\) in a coarse manner. 

\begin{figure}[t]
\centering
\includegraphics[width=0.95\linewidth]{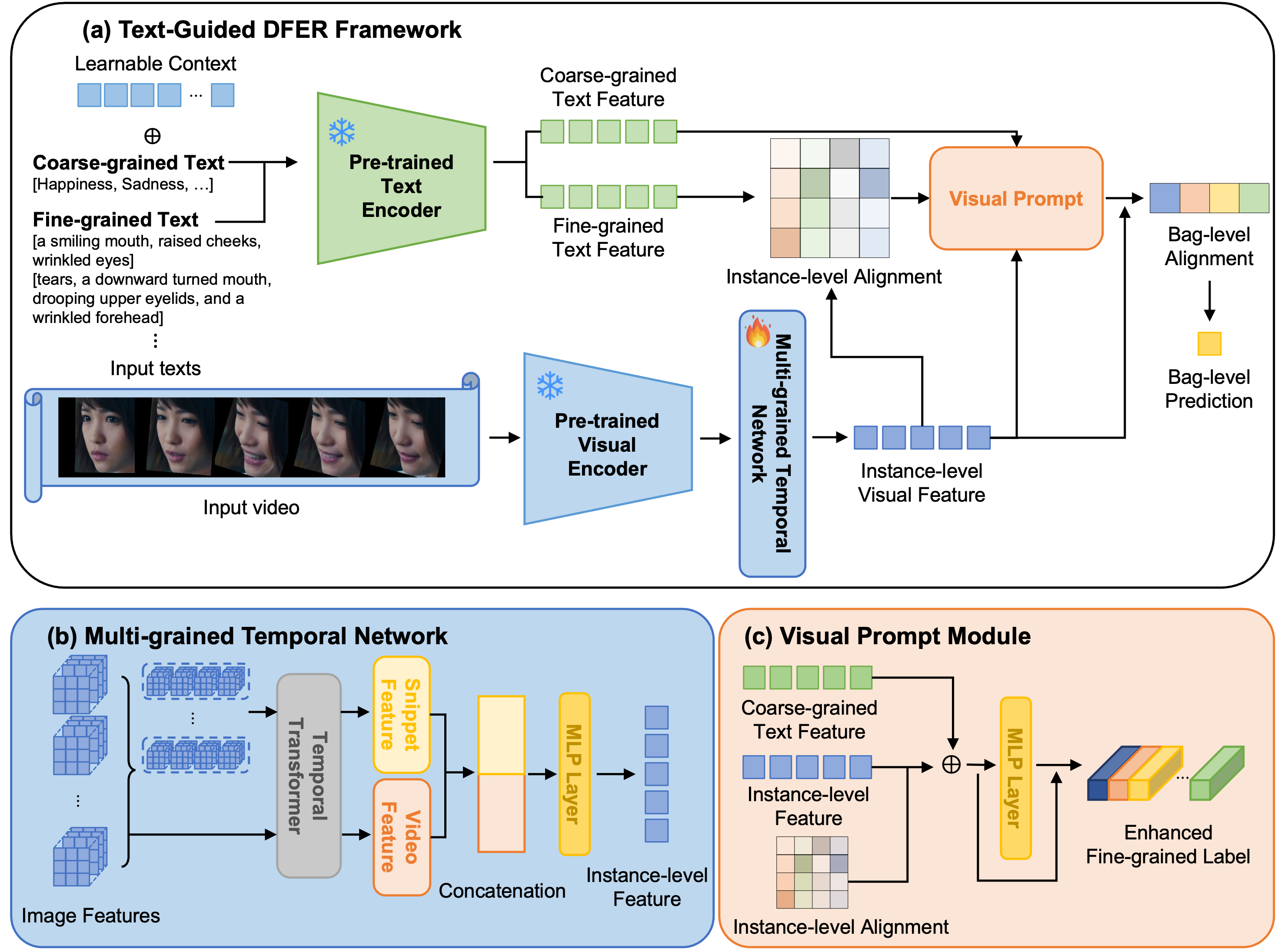}
\caption{Illustration of the proposed TG-DFER framework. (a) The overall architecture integrates visual and textual modalities within a weakly supervised learning paradigm. (b) The multi-grained temporal network captures detailed frame-wise information and long-term dynamics, ensuring temporal coherence and spatial discrimination. (c) The visual prompt module constructs enhanced fine-grained label features by enriching text labels with visual context and performing instance-level alignment.}
\label{fig_2}
\end{figure}

Previous research typically extracts video features via pretrained 3D convolutional networks (e.g., C3D \cite{bib61}, I3D \cite{bib63}), which are then classified under an MIL-based framework. Meanwhile, the large-scale vision-language pretrained model CLIP has demonstrated exceptional generalization across numerous downstream tasks and has recently shown promising results in DFER. Drawing inspiration from these successes, we propose an MIL-based approach that leverages text guidance to attenuate the influence of non-target frames, thereby enhancing performance under weak supervision. An overview of the proposed framework is demonstrated in Figure \ref{fig_2}.

\subsection{Multi-Grained Temporal Network}

To capture both short- and long-term facial expression changes, we model the temporal network to ensure local and global temporal dependencies. First, we implement a fine-grained temporal module designed to capture local temporal dependency within short video segments. This module operates on snippets extracted from the video, enabling the model to focus on the local temporal patterns. Next, we introduce a coarse-grained temporal module to capture broader temporal dependency across the entire video. This module processes the long-term frame sequence and provides a comprehensive view of the temporal dynamics. The final feature representation is obtained by combining the outputs of both the fine- and coarse-grained temporal modules.

\subsubsection{Fine-grained Temporal Module}
First, we sample \(T\) frames of the size $H \times W$ from facial video \(V\) to form an input $X \in \mathbb{R}^{T \times 3 \times H \times W}$. We extract the frame-level visual features of each video using a pre-trained CLIP image encoder, denoted as $\mathcal{F}_v$. In this paper, we employ an image encoder backbone based on ViT-B/32 \cite{bib58}, which consists of 12 layers and utilizes $32 \times 32$ patch sizes. The frame-level visual feature $X^v \in \mathbb{R}^{T \times d}$ can be formulated as
\begin{equation}
\label{eq1}
X^v = \mathcal{F}_v(X_1, X_2, X_3, \cdots, X_T),
\end{equation}
where \(d\) is the dimension size.

To capture fine-grained temporal dependency, we introduce a transform encoder layer $E_{f}$ that operates on video segments. The segments consist of frame-level features with equal-length and overlapping windows over temporal dimension. The segments are determined by window size $w$, stride $s$, and overlap frame $o$, where $s=w-o$. Overlapping ensures that each segment shares some frames with its neighboring segments, ensuring smooth transitions and continuity of context. For each feature of the segment $X^{\prime}$, we apply a transformer encoder layer, adding a learnable positional embedding $P^{f}$ to retain the temporal order within each segment: 

\begin{equation}
\label{eq3}
X^{fine} = E_{f}(X^{\prime}_0 + P^{f}_0, X^{\prime}_1 + P^{f}_1, \cdots, X^{\prime}_N + P^{f}_N),
\end{equation}
\begin{equation}
\label{eq4}
X^{\prime}_{i}=[X^v_{i\times s}, X^v_{i\times s +1}, \cdots, X^v_{i\times s+w-1}],
\end{equation}
where $N = \lfloor \frac{T-w}{s}+1 \rfloor$ is the number of segments.

\subsubsection{Coarse-grained Temporal Module}

To capture the long-range temporal dependency and coarse-grained contextual information across the entire video, we introduce a transformer encoder $E_{c}$ that processes the full sequence of the frames. The entire sequence of frame-level features $X^v$ is fed into the transformer encoder. The coarse-grained transformer encoder also adds a learnable positional embedding $P^{c}$ to maintain the temporal order within each segment, as follows:
\begin{equation}
\label{eq5}
X^{coarse} = E_{c}(X^v_0 + P^{c}_0, X^v_1 + P^{c}_1, \cdots, X^v_T + P^{c}_T).
\end{equation}

Finally, the fine- and coarse-grained temporal features are concatenated to obtain the final instance-level feature representation $X^{I} \in \mathbb{R}^{T \times d}$ as follows:
\begin{equation}
\label{eq6}
X^{I} = \mathcal{FC}([X^{fine}, X^{coarse}]),
\end{equation}
where $\mathcal{FC}(\cdot)$ is a fully-connected layer.

\subsection{Learnable Prompt}
\subsubsection{Text Prompt}
Previous studies have shown that the CLIP text encoder more effectively differentiates unique facial features when provided with detailed movement descriptions rather than single words or short phrases \cite{bib28, bib84}. Following these studies, we utilize richer textual fine-grained inputs \(y^{\prime}\) such as “a smiling mouth, widened eyes” to enhance the expressiveness of the text encoder. Nevertheless, because similar facial movements may appear across different expressions, capturing subtle differences remains challenging. To address this issue, we introduce learnable prompts \cite{bib55} that incorporate contextual information, thereby improving the model’s ability to distinguish nuanced facial expressions.

Concretely, we first transform the rich fine-grained text inputs \(y^{\prime}\) into class tokens via the CLIP tokenizer $Tok(\cdot)$, then concatenate them with a learnable context prompt $c_i \in \{c_1, \cdots, c_l\}$, where $l$ is the token size. This concatenation is given by
\begin{equation}
\label{eq2}
P_k = [c]_k^1[c]_k^2 \cdots [c]_k^M[Tok(y^{\prime})]_k,
\end{equation}
where $P_k$ denotes the tokenized text prompt at the $k$-th description. Then we leverage the frozen CLIP text encoder $\mathcal{F}_t$ to output the fine-grained text embedding feature $X^{FP} = \mathcal{F}_t(P^F) \in \mathbb{R}^{d}$, where $P^F$ is the tokenized fine text prompt with a learnable context prompt. In a similar manner, we construct coarse-grained text embeddings from categorical labels as $X^{CP} = \mathcal{F}_t(P^C) \in \mathbb{R}^{d}$, where $P^C$ denotes the tokenized coarse text prompt.

\subsubsection{Visual Prompt}

Although motion-based textual descriptions provide richer semantic cues than simple emotion labels, aligning such fine-grained text features with the corresponding visual content at the frame level remains a significant challenge. 
To address this issue, we incorporate a visual prompt mechanism \cite{bib56} that learns to identify and emphasize frames exhibiting distinctive facial movements. 
Specifically, the instance-level alignment score $A_{I}$ is computed by measuring the similarity between each frame's visual feature $X^{I}$ and the fine-grained textual feature $X^{FP}$.
This score provides a set of attention weights that determine the contribution of each frame. 
The visual prompt $V^{P}$  is subsequently generated by taking a weighted sum of the frame-level visual features, which distills the visual information into a single vector representing the most relevant context as follows:

\begin{equation}
\label{eq7}
V^P = \mathcal{N}(A_I\cdot X^{I}),
\end{equation}
\begin{equation}
A_I = \mathcal{M}(X^{I}, X^{FP})/\tau,
\end{equation}
where $\mathcal{N}$ is the normalization, $\mathcal{M}(\cdot)$ denotes cosine similarity and $\tau$ is a temperature scaling parameter. To enhance semantic representation, we refine the fine-grained label feature by incorporating both the visual prompt and the coarse-grained textual embedding. We first compute an intermediate feature $\Hat{X}^p$ using element-wise addition between the fine-grained text prompt $X^{FP}$ and the visual prompt $V^P$,
\begin{equation}
\label{eq9}
\Hat{X}^p=X^{FP} \oplus V^P,
\end{equation}
where $\oplus$ denotes element-wise addition. 
This operation serves to inject the visual context from $V^{P}$ into the textual feature. 
The subsequent MLP layer acts as a non-linear fusion module, which is tasked with learning the intricate relationships between the visual and textual features to create a unified and effective representation.
This design is analogous to residual learning, allowing for the efficient integration of new information while preserving the original feature representation.

Then, we apply the multi-layer perceptron (MLP) and fuse the result with the coarse-grained text embeddings $X^{CP}$,
\begin{equation}
\label{eq10}
\Tilde{X}^p = MLP(\Hat{X}^p) + X^{CP}.
\end{equation}
This two-stage fusion process, which involves injecting visual context via addition and then modeling their complex interaction with an MLP, allows the final label representation to benefit from both fine-grained contextual alignment and high-level categorical guidance.

This approach enables enhanced fine-grained label feature to retrieve the relevant visual context from videos. Finally, we can obtain bag-level prediction $\Tilde{y}_{k}$ with respect to the $k^{th}$ class as follows:
\begin{equation}
\label{eq12}
\Tilde{y}_{k} = \frac{exp(\mathcal{M}(X^{I}, \Tilde{X}^{p}_{k})/\tau)}{\sum^C_{i=1}exp(\mathcal{M}(X^{I}, \Tilde{X}^{p}_{i})/\tau)},
\end{equation}
where $C$ is the number of classes and $\tau$ is the temperature hyper-parameter for scaling. 

During training, the CLIP text and image encoders remain frozen to leverage their powerful generalization capabilities while ensuring parameter efficiency.
Our proposed multi-grained temporal network is trained from scratch in an end-to-end manner.

Following prior work \cite{bib24}, we adopt MIL mechanism by employing the cross-entropy loss $\mathcal{L}_{nce}$ between the bag-level expression prediction score and the corresponding ground truth.

\section{Experiments}

We conducted extensive experiments on two widely used DFER datasets, namely DFEW \cite{bib59} and FERV39k \cite{bib60}. This section introduces the datasets and details their implementation specifications. Next, we evaluate the effectiveness of each component of our methodology using the DFEW dataset. Subsequently, we compare our proposed method with several state-of-the-art approaches, highlighting its efficacy of our approach through various visualizations. 

\subsection{Datasets}

\textbf{DFEW} \cite{bib59} dataset comprises over 16,000 video clips, making it the largest benchmark for DFER in uncontrolled environments. These clips are sourced from thousands of movies globally and include various practical challenges, such as extreme illuminations, self-occlusions, and varying head poses. Each video clip is annotated by several independent individuals under professional supervision and classified into one of seven basic emotions: \textit{Happiness}, \textit{Sadness}, \textit{Neutral}, \textit{Anger}, \textit{Surprise}, \textit{Disgust}, and \textit{Fear}. The dataset is divided into five equal, non-overlapping parts (fd1-fd5). To ensure fair comparisons among different methods, we utilize the 5-fold cross-validation setting provided by DFEW.

\textbf{FERV39k} \cite{bib60}  dataset comprises 38,935 video clips captured in-the-wild environments, making it the largest of its type for DFER. These clips are sourced from four broad scenarios, which are further categorized into 22 detailed contexts: crime, daily life, speech, and war. This dataset is pioneering due to its large scale, scenario-based segmentation, and cross-domain applicability. Each video clip is annotated by 30 individuals and classified into one of the seven basic expressions, similar to the DFEW dataset. To ensure fair comparisons, the dataset is randomly divided into training and testing sets, with no overlap between them.

\subsection{Implementation Details}

In our experiments, we resize all images to 224 $\times$ 224 pixels and extract a total of 16 frames from each video, followed by a sampling strategy as in previous works \cite{bib27}, \cite{bib28}, \cite{bib24}, \cite{bib25}. To mitigate overfitting, we employ various data augmentation techniques, including random cropping, rotation, horizontal flipping, and color jittering. For all datasets, we utilize the ViT-B/32 \cite{bib58} of CLIP as the backbone model. The maximum number of text tokens is set to 77, as per the official CLIP specifications, and the temperature hyperparameter \( \tau \) is fixed at 0.01. The framework is implemented using a PyTorch-GPU and trained on an RTX A6000 GPU. The parameters are optimized with a mini-batch size of 48 using the SGD optimizer. The initial learning rates is set to 1e-5 for the CLIP image encoder, 1e-3 for the learnable prompts, and 1e-2 for the remaining parameters, except for the CLIP text encoder. We employ a MultiStepLR scheduler with milestones at epochs 30 and 40, applying a gamma value of 0.1 to adjust the learning rate.

Following prior methods, we employ two evaluation metrics: the weighted average recall (WAR) and unweighted average recall (UAR). The WAR measures the mean accuracy by considering the proportion of instances per class, whereas the UAR calculates the mean accuracy per class divided by the number of classes without accounting for the number of instances per class. The WAR is deemed a critical metric for evaluation.

\subsection{Comparison with State-of-the-Art Methods}

In this section, we compare our results with those of several state-of-the-art methods on the DFEW and FERV39k benchmarks. Following previous methods \cite{bib28}, \cite{bib41}, \cite{bib24}, \cite{bib23}, \cite{bib25}, the DFEW experiments are conducted with 5-fold cross-validation, while the FERV39k experiments are conducted with distinct training and test sets. 

\begin{sidewaystable}[!htbp]
\caption{Comparison with the state-of-the-art methods on DFEW.}
\label{tab5}
\begin{adjustbox}{scale=1,center}
\begin{tabular}{clll|cllllll|cc}

\toprule
\multicolumn{4}{c|}{\multirow{2}{*}{Method}} & \multicolumn{7}{c|}{Accuracy of Each Facial Expression} & \multicolumn{2}{c}{Metrics (\%)} \\ \cline{5-11} \cline{12-13} 
\multicolumn{4}{c|}{}               & Happiness & Sadness & Neutral & Anger & Surprise & Disgust & Fear & UAR & WAR  \\ \hline
\multicolumn{4}{c|}{C3D \cite{bib61}}            & 75.17 & 39.49 & 55.11 & 62.49 & 45.00 & 1.38 & 20.51 & 42.74 & 53.54   \\ 
\multicolumn{4}{c|}{P3D \cite{bib62}}            & 74.85 & 43.40 & 54.18 & 60.42 & 50.99 & 0.69 & 23.28 & 43.97 & 54.47  \\ 
\multicolumn{4}{c|}{I3D-RGB \cite{bib63}}        & 78.61 & 44.19 & 56.69 & 55.87 & 45.88 & 2.07 & 20.51 & 43.40 & 54.27 \\ 
\multicolumn{4}{c|}{3D ResNet18 \cite{bib64}}    & 73.13 & 48.26 & 50.51 & 64.75 & 50.10 & 0.00 & 26.39 & 44.73 & 54.98 \\ 
\multicolumn{4}{c|}{R(2+1)D18 \cite{bib65}}      & 79.67 & 39.07 & 57.66 & 50.39 & 48.26 & 3.45 & 21.06 & 42.79 & 53.22 \\ 
\multicolumn{4}{c|}{ResNet18-LSTM \cite{bib59}}  & 83.56 & 61.56 & 68.27 & 65.29 & 51.26 & 0.00 & 29.34 & 51.32 & 63.85 \\ 
\multicolumn{4}{c|}{EC-STFL \cite{bib59}}        & 79.18 & 49.05 & 57.85 & 60.98 & 46.15 & 2.76 & 21.51 & 45.35 & 56.51 \\ 
\multicolumn{4}{c|}{Former-DFER \cite{bib9}}    & 84.05 & 62.57 & 67.52 & 70.03 & 56.43 & 3.45 & 31.78 & 53.69 & 65.70 \\ 
\multicolumn{4}{c|}{NR-DFERNet \cite{bib23}}     & 88.47 & 64.84 & 70.03 & 75.09 & 61.60 & 0.00 & 19.43 & 54.21 & 68.19 \\
\multicolumn{4}{c|}{DPCNet \cite{bib11}}         & 89.93 & 64.61 & 67.12 & 63.18 & 53.67 & 15.86 & 31.56 & 57.11 & 66.32 \\
\multicolumn{4}{c|}{EST \cite{bib22}}            & 86.87 & 66.58 & 67.18 & 71.84 & 47.52 & 5.52 & 28.49 & 53.94 & 65.85 \\
\multicolumn{4}{c|}{LOGO-Former \cite{bib66}}    & 85.39 & 66.52 & 68.94 & 71.33 & 54.59 & 0.00 & 32.71 & 54.21 & 66.98 \\
\multicolumn{4}{c|}{AEN \cite{bib67}}            & 89.24 & 69.38 & 70.67 & 72.08 & 59.07 & 4.17 & 32.00 & 56.66 & 69.37 \\
\multicolumn{4}{c|}{IAL \cite{bib25}}            & 87.95 & 67.21 & 70.10 & 76.06 & 62.22 & 0.00 & 26.44 & 55.71 & 69.24 \\
\multicolumn{4}{c|}{MIDAS \cite{bib41}}            & 87.40 & 67.34 & 58.64 & 68.06 & 59.65 & 28.69 & 44.50 & 57.45 & 69.16 \\
\multicolumn{4}{c|}{M3DFEL \cite{bib24}}         & 89.59 & 68.38 & 67.88 & 74.24 & 59.69 & 0.00 & 31.63 & 56.10 & 69.25 \\
\multicolumn{4}{c|}{CLIPER \cite{bib27}}         & - & - & -     & -     & -     & -    & -     & 57.56 & 70.84 \\
\multicolumn{4}{c|}{DFER-CLIP \cite{bib28}}      & 91.12 & 75.34 & 71.15 & 74.09 & 56.30 & 11.72 & 37.81 & 59.61 & 71.25 \\
\hline
\multicolumn{4}{c|}{TG-DFER (Ours)}     & 90.92 & 74.92 & 71.41 & 73.36 & 59.43 & 13.10 & 38.03 & \textbf{60.17} & \textbf{71.62} \\
\bottomrule

\end{tabular}
\end{adjustbox}
\end{sidewaystable}

As shown in Table \ref{tab5}, the proposed TG-DFER outperforms the compared methods with a UAR of 60.17\% and a WAR of 71.62\%, establishing a new state-of-the-art. Specifically, our approach surpasses the MIL-based M3DFEL \cite{bib24} by 4.07\% in UAR and 2.37\% in WAR, and exceeds DFER-CLIP \cite{bib28}, which incorporates text information, by 0.56\% in UAR and 0.37\% in WAR. Notably, our method achieves the highest accuracy on \textit{Neutral} expression, which is often the most ambiguous, and also performs strongly on \textit{Disgust} and \textit{Fear}, both of which are underrepresented in the datset. In the DFEW dataset, \textit{Disgust} and \textit{Fear} sequences constitute only 1.22\% and 8.14\% of the data, respectively \cite{bib23}.

As illustrated in Table \ref{tab6}, our method surpasses several state-of-the-art methods on the FERV39k benchmark, a more challenging DFER dataset that generally produces lower overall accuracy than DFEW. Concretely, TG-DFER achieves a UAR of 41.50\% and a WAR of 51.67\%. TG-DFER outperforms M3DFEL \cite{bib24} by 5.56\% in UAR, 4.00\% in WAR, and exceeds DFER-CLIP \cite{bib28} by 0.23\% in UAR, 0.02\% in WAR, respectively. These results demonstrate that combining a text-guided MIL paradigm improves performance beyond existing state-of-the-art methods, confirming the effectiveness of our proposed approach in addressing the inherent challenges of in-the-wild DFER.

\begin{table}[!t]
\begin{center}
\caption{Comparison with the state-of-the-art methods on FERV39k.}
\label{tab6}
\begin{tabular}{clll|cc}

\toprule
\multicolumn{4}{c|}{\multirow{2}{*}{Method}} &   \multicolumn{2}{c}{Metrics (\%)} \\ 
\cline{5-6} 
\multicolumn{4}{c|}{}               & UAR & WAR  \\ \hline
\multicolumn{4}{c|}{C3D \cite{bib61}}            & 22.68 & 31.69 \\ 
\multicolumn{4}{c|}{P3D \cite{bib62}}            & 23.20 & 33.39 \\ 
\multicolumn{4}{c|}{I3D-RGB \cite{bib64}}        & 30.17 & 38.78 \\ 
\multicolumn{4}{c|}{3D ResNet18 \cite{bib64}}    & 26.67 & 37.57   \\ 
\multicolumn{4}{c|}{R(2+1)D18 \cite{bib65}}      & 31.55 & 41.28 \\ 
\multicolumn{4}{c|}{ResNet18-LSTM \cite{bib59}}  & 30.92 & 42.95  \\ 
\multicolumn{4}{c|}{ResNet18-ViT}                & 38.35 & 48.43  \\ 
\multicolumn{4}{c|}{Former-DFER \cite{bib9}}     & 37.20 & 46.85\\ 
\multicolumn{4}{c|}{NR-DFERNet \cite{bib23}}     & 33.99 & 45.97\\ 
\multicolumn{4}{c|}{LOGO-Former \cite{bib66}}    & 38.22 & 48.13  \\ 
\multicolumn{4}{c|}{IAL \cite{bib25}}            & 35.82 & 48.54 \\ 
\multicolumn{4}{c|}{MIDAS \cite{bib41}}          & 39.20 & 47.37 \\ 
\multicolumn{4}{c|}{AEN \cite{bib67}}            & 38.18 & 47.88   \\ 
\multicolumn{4}{c|}{M3DFEL \cite{bib24}}         & 35.94 & 47.67   \\ 
\multicolumn{4}{c|}{CLIPER \cite{bib27}}         & 41.23 & 51.34   \\ 
\multicolumn{4}{c|}{DFER-CLIP \cite{bib28}}      & 41.27 & 51.65   \\ 
\hline
\multicolumn{4}{c|}{TG-DFER (Ours)}                        & \textbf{41.50} & \textbf{51.67}  \\ 
\bottomrule

\end{tabular}
\end{center}
\end{table}

\subsection{Ablation Studies} \label{ablation}
To assess the effectiveness of each component in our approach, we perform a component ablation analysis on the DFEW benchmark using 5-fold cross-validation. Specifically, we explore the multi-grained temporal module under different temporal depths and analyze the impact of hyper-parameters in the fine-grained temporal module. We then investigate the influence of visual prompt placement across various text prompts. Finally, we examine the contribution of fine-grained label features, which exert a significant influence on the target expression.

\begin{table}[!t]
\begin{center}
\caption{Comparison of our multi-grained temporal network with different depths.}
\label{tab1}
\begin{tabular}{cl|cc|cc}

\toprule
\multicolumn{2}{c|}{\multirow{2}{*}{Setting}} &  \multicolumn{2}{c|}{Method} & \multicolumn{2}{c}{Metrics (\%)}\\ 
\cline{3-4} \cline{5-6}
\multicolumn{2}{c|}{}             & Fine- & Coarse- & UAR & WAR \\ \hline
\multicolumn{2}{c|}{a}            & \ding{55} & \ding{55} & 57.66 & 69.14 \\ 
\multicolumn{2}{c|}{b}            & \ding{55} & 1 & 59.83 & 71.34 \\ 
\multicolumn{2}{c|}{c}            & \ding{55} & 2 & 59.45 & 71.18 \\  \hline 
\multicolumn{2}{c|}{d}            & 1 & \ding{55} & 58.50 & 69.24 \\  
\multicolumn{2}{c|}{e}            & 1 & 1 & 60.17 & \textbf{71.62} \\  
\multicolumn{2}{c|}{f}            & 1 & 2 & \textbf{60.58} & 71.46 \\  \hline 
\multicolumn{2}{c|}{g}            & 2 & \ding{55} & 57.73 & 69.27 \\ 
\multicolumn{2}{c|}{h}            & 2 & 1 & 60.40 & 71.51 \\  
\multicolumn{2}{c|}{i}            & 2 & 2 & 60.27 & 71.51 \\  \hline 
\multicolumn{2}{c|}{{j}}            & {1} & {7}& {59.10} & {70.93} \\  
\multicolumn{2}{c|}{{k}}            & {7} & {1} & {59.24} & {70.84} \\  
\multicolumn{2}{c|}{{l}}            & {7} & {7} & {59.29} & {70.92} \\
\bottomrule

\end{tabular}
\end{center}
\end{table}

\begin{table}[!t]
\begin{center}
\caption{Evaluation of the stride and window width in the fine-grained temporal network.}
\label{tab2}
\begin{tabular}{cc|cc}

\toprule
\multicolumn{2}{c|}{Setting} & \multicolumn{2}{c}{Metrics (\%)} \\ \hline
Stride & Window Width & UAR & WAR  \\ \hline
\ding{55} & 2 & 59.58 & 71.39 \\ 
\ding{55} & 4 & 59.58 & 71.32 \\ 
\ding{55} & 8 & 59.50 & 71.43 \\ \hline
1 & 2 & 59.59 & 71.19 \\ 
1 & 4 & \textbf{60.17} & \textbf{71.62} \\ 
1 & 8 & 59.83 & 71.51 \\ \hline
2 & 4 & 60.09 & 71.48 \\ 
2 & 8 & 59.76 & 71.51 \\ 
3 & 4 & 59.81 & 71.48 \\ 
\bottomrule

\end{tabular}
\end{center}
\end{table}

\subsubsection{Evaluation of Multi-Grained Temporal Module with Different Depths}
First, We conduct an ablation study on multi-grained temporal networks. As shown in Table \ref{tab1}, we analyze the effects of fine- and coarse-grained temporal networks along with different temporal depths. In settings (d, g) that utilize only local features, the average UAR is 58.12\% and the average WAR is 69.26\%. In contrast, the global-only approach (b, c) achieve an average UAR of 59.64\% and an average WAR of 71.26\%. Notably, both networks show performance degradation as the temporal depth increases. Based on the UAR, the optimal depth for the fine-grained network is 1, whereas that for the coarse-grained network is 2. In terms of the WAR, the fine-grained network performs best at a depth of 1, whereas the optimal depth for the coarse networks is also 1. Although deeper models typically yield higher performance, in our case, the performance is lower. 
{This is attributed to a mismatch between the model's increased complexity and the low inherent temporal complexity of the DFER datasets.
The key emotional cues in these datasets are often concentrated in short, decisive moments, meaning a shallow network is sufficient. 
An overly deep network may overfit to sparse temporal cues for the simpler data, disrupting the delicate balance between feature scales by granting undue importance to the broader, coarse-grained context over the critical fine-grained nuances. 
This issue is exacerbated because the temporal model is trained from scratch and is thus more susceptible to poor generalization on these patterns. 
Fine-grained networks focus on short-term, local features to capture subtle expression dynamics, while coarse-grained networks incorporate broader temporal context. 
Consequently, leveraging both feature scales in tandem at a shallow depth yields superior performance by effectively balancing short-term nuances with overarching temporal cues.}

\subsubsection{Evaluation of Hyper-Parameters in the Fine-grained Temporal Module} 
We conduct an ablation study on the key hyper-parameters in the fine-grained temporal network. As shown in Table \ref{tab2}, the performance of the fine-grained temporal network is evaluated by varying the stride and window width parameters. In the case without stride, changing the window width to 2, 4, or 8 result in minimal performance changes, with UAR values of 59.58\%, 59.58\%, and 59.50\%, and WAR values of 71.39\%, 71.32\%, and 71.43\%, respectively. Conversely, when the stride is set to 1, the highest performance is observed with a window width of 4, achieving a UAR of 60.17\% and a WAR of 71.62\%. The results indicate that smaller window widths, corresponding to shorter snippets, are associated with increased UAR and WAR. Dividing the video into smaller segments allows for fine-grained local expression information to be represented, thereby better capturing subtle changes in the expression. Additionally, larger strides, which involve overlapping frames, generally perform better, although the performance is contingent upon the specific combination of the stride and window width. This demonstrates the importance of balancing the stride and window width to effectively capture temporal patterns in the data. When the stride is increased to 2, the performance remained high with a window width of 4 (UAR 60.09\%, WAR 71.48\%), but showed a slight decrease with a window width of 8 (UAR 59.76\%, WAR 71.51\%). With a stride of 3 and a window width of 4, the performance exhibited little change, recording a UAR of 59.81\% and a WAR of 71.48\%. This underscores the critical role of hyper-parameter tuning in a fine-grained temporal network, demonstrating that an appropriate combination of stride and window width significantly enhances the performance.

\begin{table}[!t]
\begin{center}
\caption{The ablation study of visual prompt location with different text prompts.}
\label{tab4}
\begin{tabular}{cl|cc|cc}

\toprule
\multicolumn{2}{c|}{\multirow{2}{*}{Setting}} &  \multicolumn{2}{c|}{Prompts} & \multicolumn{2}{c}{Metrics (\%)}\\ 
\cline{3-4} \cline{5-6}
\multicolumn{2}{c|}{}             & Text Prompt & Visual Prompt & UAR & WAR \\ \hline
\multicolumn{2}{c|}{a}            & Class & \ding{55} & 59.12 & 71.40 \\ 
\multicolumn{2}{c|}{b}            & Class & Add & 60.07 & 71.58 \\ 
\multicolumn{2}{c|}{c}            & Class & Prepend & 59.58 & 71.49 \\ \hline 
\multicolumn{2}{c|}{d}            & Descriptors & \ding{55} & 59.12 & 71.34 \\ 
\multicolumn{2}{c|}{e}            & Descriptors & Add & 60.23 & 71.54 \\ 
\multicolumn{2}{c|}{f}            & Descriptors & Prepend & 59.14 & 71.43 \\ \hline 
\multicolumn{2}{c|}{g}            & [Learnable] Class & \ding{55} & 59.46 & 71.07 \\ 
\multicolumn{2}{c|}{h}            & [Learnable] Class & Add & 60.17 & \textbf{71.62} \\ 
\multicolumn{2}{c|}{i}            & [Learnable] Class & Prepend & \textbf{60.36} & 71.47 \\ 
\multicolumn{2}{c|}{j}            & [Learnable] Descriptors & \ding{55} & 58.56 & 71.17 \\ 
\multicolumn{2}{c|}{k}            & [Learnable] Descriptors & Add & 59.23 & 71.32 \\ 
\multicolumn{2}{c|}{l}            & [Learnable] Descriptors & Prepend & 59.75 & \textbf{71.62} \\ 
\bottomrule

\end{tabular}
\end{center}
\end{table}

\subsubsection{Evaluation of Visual Prompt Location with Different Text Prompts}
We analyze the impact of different prompts and visual prompt on the performance of the adaptive instance aggregation module in Table \ref{tab4}. The text prompts are configured by adding ``Class,'' ``Descriptors,'' and learnable prompts according to the baseline method \cite{bib28}. When the text prompt is set to ``Class'' and the visual prompt is ``Add'' or ``Prepend,'' there is a noticeable improvement in performance compared to when the visual prompt is excluded. For instance, with the ``Class'' prompt and visual prompt ``Add,'' the UAR is 60.07\% and the WAR is 71.58\%. Similarly, the ``Descriptors'' prompt shows improved performance with visual prompt ``Add'' or ``Prepend.'' Specifically, the ``Descriptors'' prompt with visual prompt ``Add'' results in a UAR of 60.23\% and a WAR of 71.54\%. Notably, incorporating learnable prompts significantly boosts the performance. For example, the learnable ``Class'' prompts with visual prompt ``Add'' or ``Prepend'' achieve UARs of 60.17\% and 60.36\%, and WARs of 71.62\% and 71.47\%, respectively. The learnable ``Descriptors'' prompt with visual prompt ``Prepend'' achieves a UAR of 59.75\% and a WAR of 71.62\%. 

\begin{table}[!t]
\begin{center}
\caption{Evaluation of enhanced fine-grained label for Target Expression.}
\label{tab_plus}
\begin{tabular}{cc|cc}

\toprule
\multicolumn{2}{c|}{Setting} & \multicolumn{2}{c}{Metrics (\%)} \\ \hline
Influence & \textit{\#} of frames & UAR & WAR  \\ \hline
Highest & 1 & 59.97 & 71.12 \\ 
Highest & 2 & \textbf{60.09} & \textbf{71.27} \\ 
Highest & 3 & 59.74 & 71.12 \\ \hline
Lowest & 1 & 55.41 & 67.39 \\ 
Lowest & 2 & 57.18 & 68.88 \\ 
Lowest & 3 & 58.00 & 69.63 \\ 
\bottomrule

\end{tabular}
\end{center}
\end{table}

\subsubsection{Evaluation of Enhanced Fine-grained label for Target Expression}
\label{influence}
We evaluate the effectiveness of the fine-grained label feature for identifying the target facial expression by analyzing inference results using frames selected based on the ``Highest'' and ``Lowest'' influence of this feature. As shown in Table \ref{tab_plus}, inference with frames exhibiting the ``Highest'' fine-grained feature influence significantly outperforms inference using frames with the ``Lowest'' influence. Specifically, selecting two frames with the ``Highest'' influence achieves optimal performance (60.09\% UAR and 71.27\% WAR), indicating that frames with strong fine-grained feature influence reliably represent the target emotion. Conversely, inference using the frames with the ``Lowest'' influence results in considerably lower accuracy (55.41\% UAR and 67.39\% WAR when using one frame), underscoring the limited representational capability of frames with low fine-grained feature influence. These results confirm that the influence of fine-grained label features plays a critical role in accurately recognizing target expressions.

\subsection{Qualitative Analyses}
We conduct qualitative analyses to demonstrate the effectiveness of our method further. 

\begin{figure*}[!t]
\begin{center}
\resizebox{\linewidth}{!}
{
    \begin{tabular}{c @{\;\;} c @{\;\;} c}
    \includegraphics[width=\linewidth]{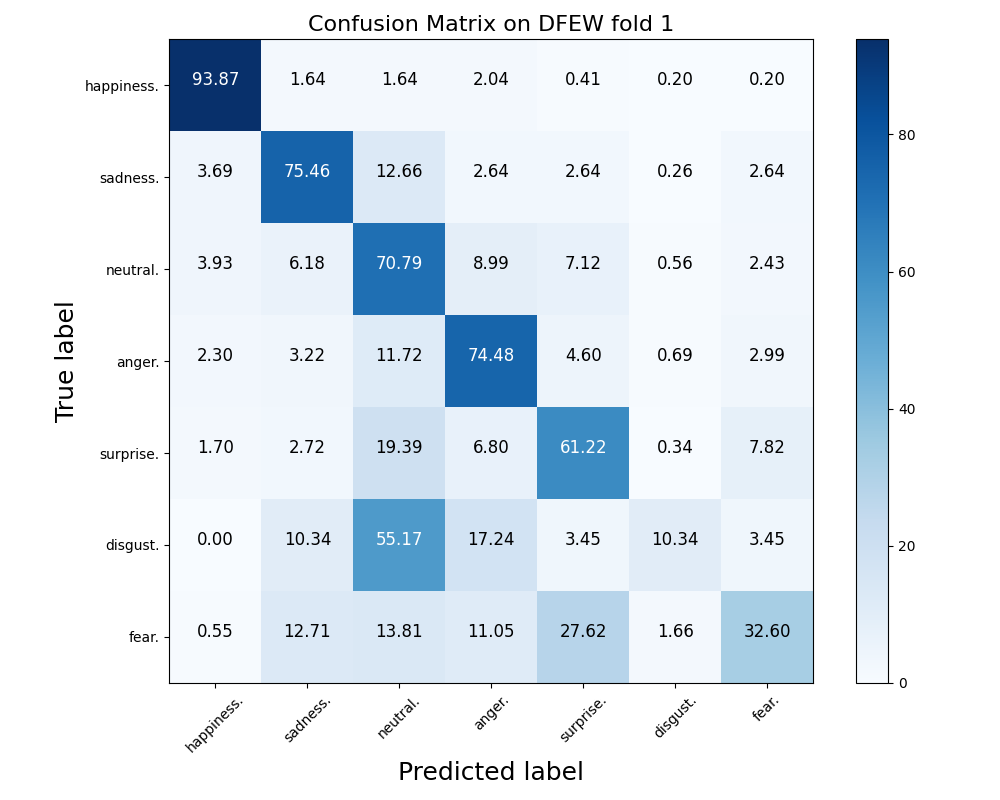} &
    \includegraphics[width=\linewidth]{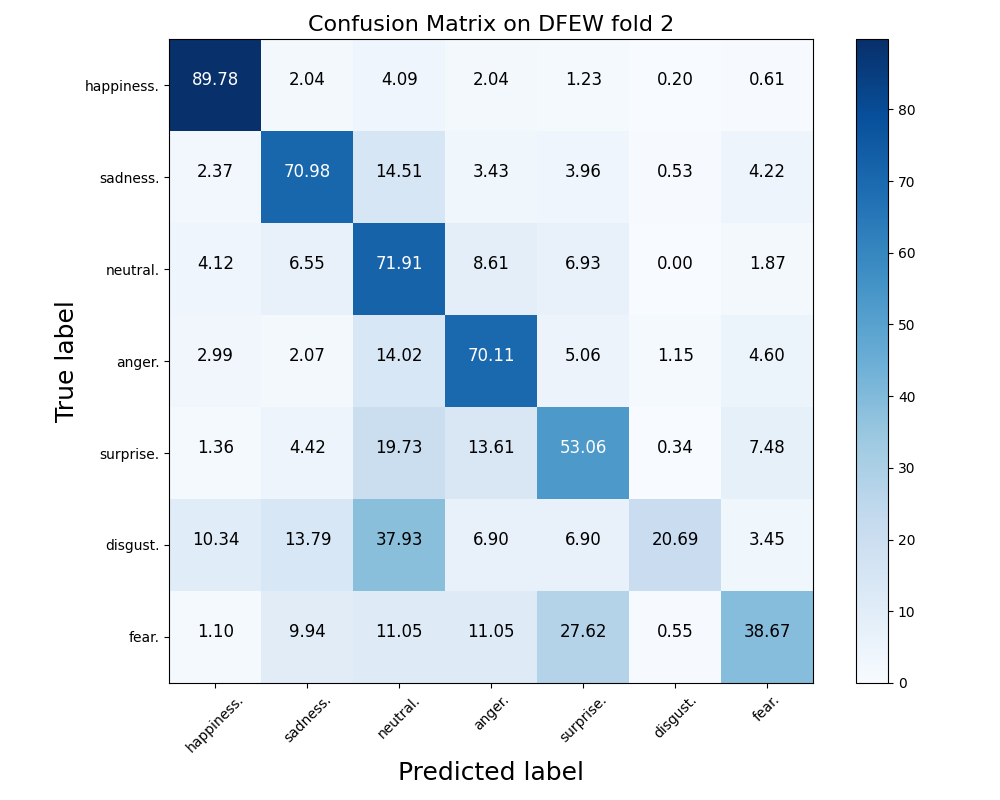} &
    \includegraphics[width=\linewidth]{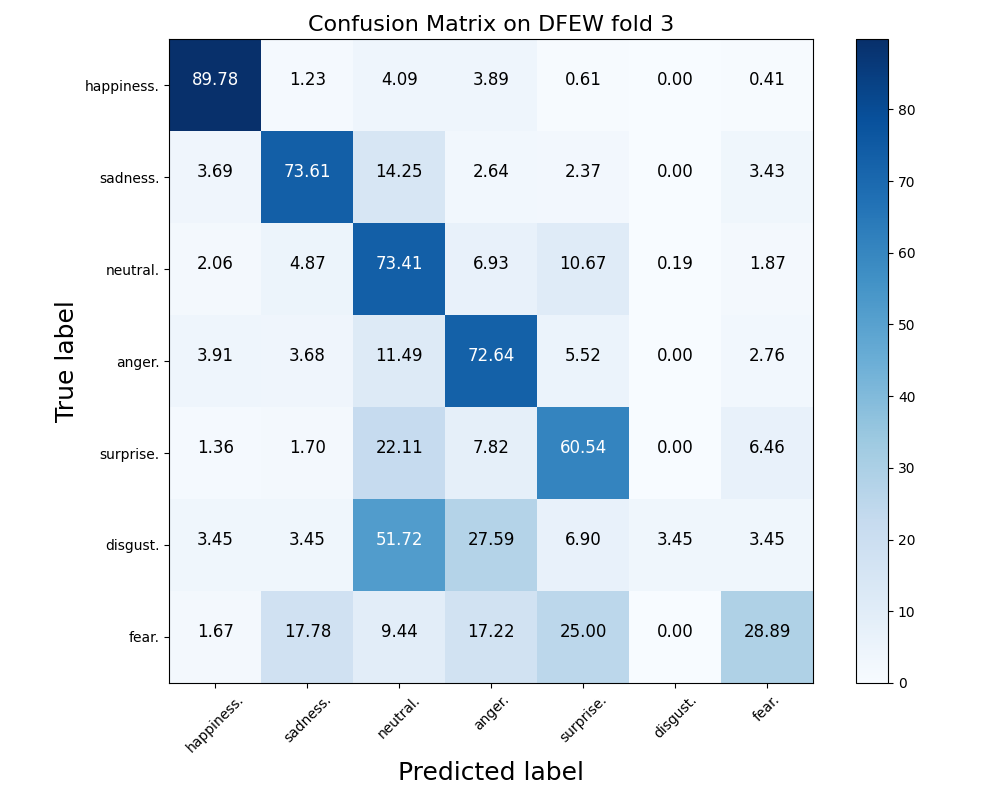} \\
    \Huge{(a)} & \Huge{(b)} & \Huge{(c)} \\
    \includegraphics[width=\linewidth]{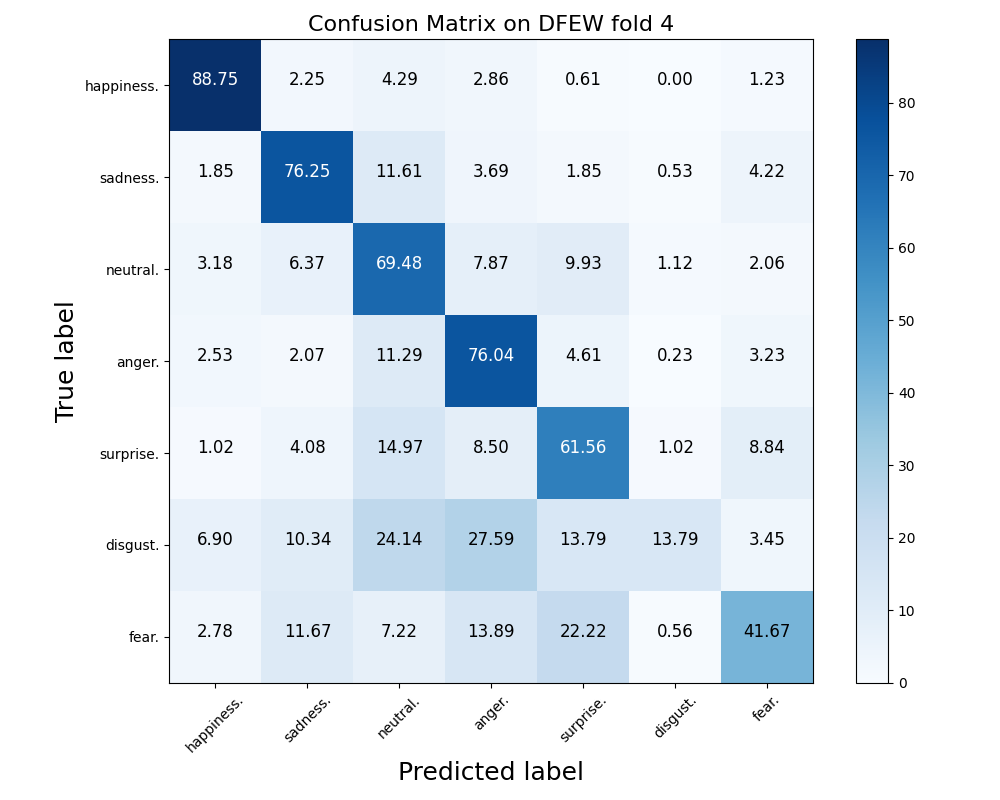} &
    \includegraphics[width=\linewidth]{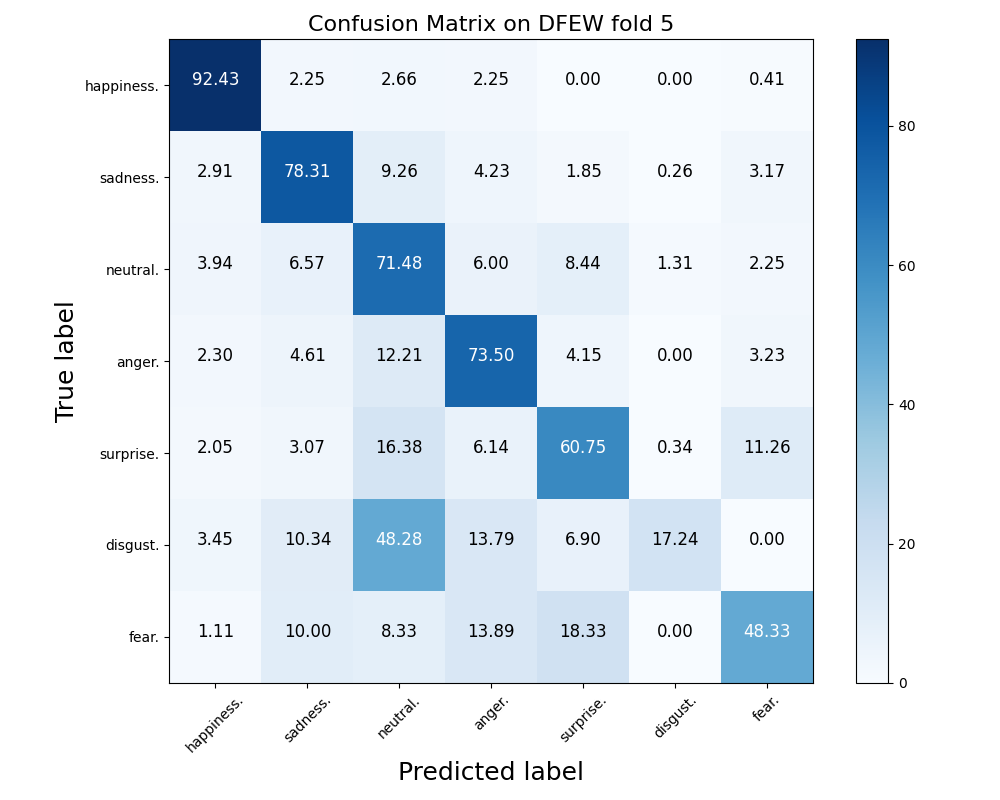} &
    \includegraphics[width=\linewidth]{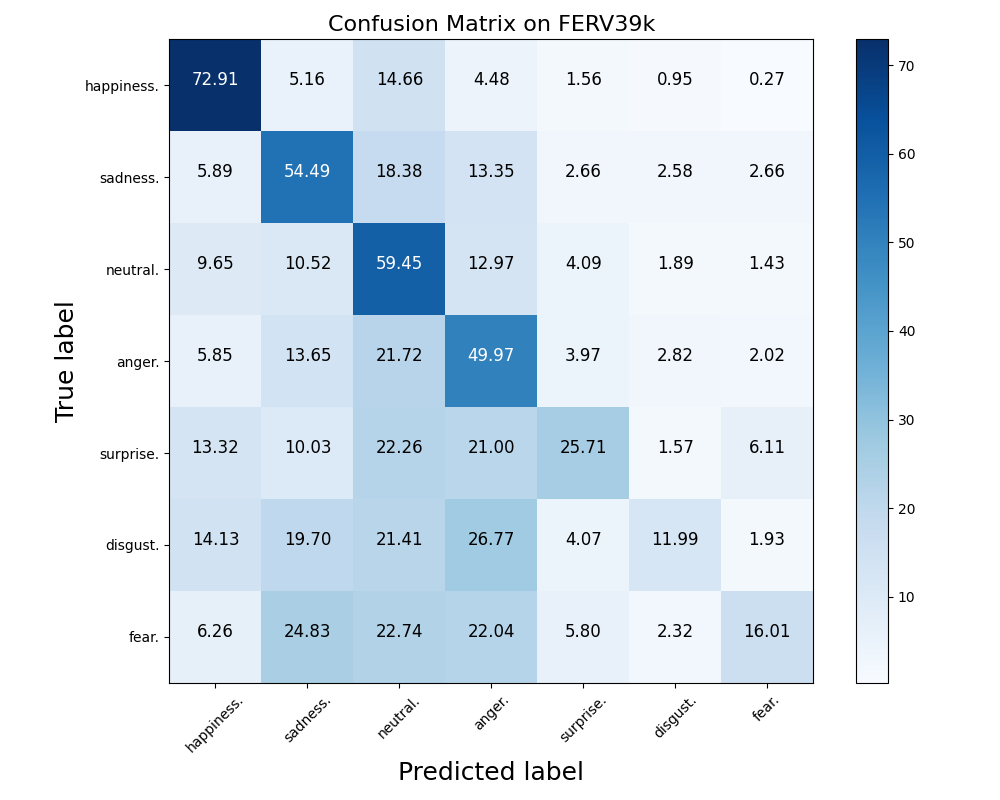} \\ 
    \Huge{(d)} & \Huge{(e)} & \Huge{(f)}
    \end{tabular}
}
\end{center}
\caption{Confusion matrix of our proposed TG-DFER evaluated on 5-fold DFEW (a)-(e) and FERV39k (f).}
\label{fig_4}
\end{figure*}

\subsubsection{Confusion Matrix} 
{
We visualize the confusion matrix of our method evaluated on 5 fold of the DFEW dataset (Fig. 3 (a)-(e)) and on the FERV39k dataset (Fig. 3 (f)). 
Across both benchmarks, the model struggles with underrepresented expressions like \textit{Disgust} and \textit{Fear}.
This difficulty is a known issue stemming from the severe data imbalance in these datasets, which results in insufficient training examples for these classes} \cite{bib23, bib24}{. 
Conversely, the model demonstrates strong predictive accuracy for expressions such as \textit{Happiness} and, notably, \textit{Neutral}.
The \textit{Neutral} expression is inherently ambiguous and frequently shares visual cues with other emotions. 
Rather than this ambiguity leading to poor performance, our model appears to learn robust features that effectively disentangle it from other expressions. 
This suggests that the model becomes adept at navigating the subtleties of ambiguous facial states.
Overall, these results indicate that while the model's performance is impacted by extreme data imbalance, it effectively handles the inherent ambiguity between expressions, demonstrating robust recognition capabilities in challenging in-the-wild scenarios.
}
\begin{figure}[t]
\centering
\includegraphics[width=0.95\linewidth]{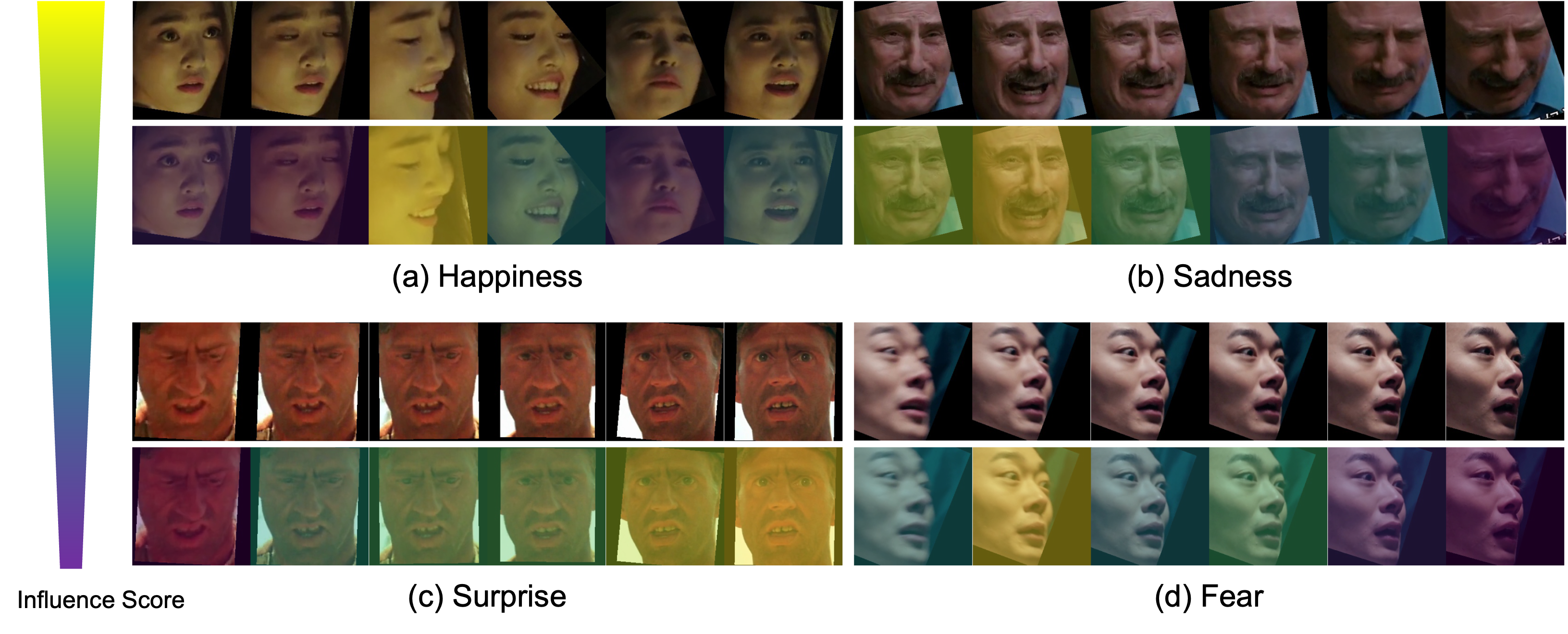}
\caption{Visualization results of enhanced fine-grained label feature influence in the DFEW dataset.}
\label{fig_5}
\end{figure}

\subsubsection{Visualization of Enhanced Fine-grained Label Feature Influence}
We visualize the influence of the fine-grained label feature as described in Section \ref{influence}. In our model, we utilize the influence of the fine-grained label feature $\Tilde{X}^p$ after normalization. Specifically, we apply min-max normalization exclusively for visualization of frame-level influence, not during training. This normalization can be formulated as follows:

\begin{equation}
\label{eq13}
\Tilde{X}^p_{norm} = \frac{\Tilde{X}^p - \Tilde{X}^p_{min}}{\Tilde{X}^p_{max} - \Tilde{X}^p_{min}}.
\end{equation} 

The color bar, ranging from purple to yellow, indicates the level of temporal attention, with dark purple representing low attention and light yellow representing high attention. Fig. \ref{fig_5} (a) depicts a video sample related to \textit{Happiness}, with other expressions also present in the video. Our model demonstrates that the first and second frames, which are not associated with \textit{Happiness}, have low influence. In contrast, the third frame, which is closely related to \textit{Happiness}, exhibits the highest influence, and frames generally related to \textit{Happiness} have the highest influence. Fig. \ref{fig_5} (b) presents a video sample related to \textit{Sadness}, where the same expression is present throughout the video, but the intensity of the expression varies. Our model indicates that second frame, which is a good representation of \textit{Sadness}, has a higher influence than sixth frame, which is a noisy frame with a leaning face. Fig. \ref{fig_5} (c) and (d) also show the frame influence of different emotional intensities on the expressions of \textit{Surprise} and \textit{Fear}. This demonstrates that our model recognizes that frames unrelated to the target expression have low impact, whereas frames related to the target expression have high impact. 

\section{Discussion}

In this study, we proposed and validated a novel framework, TG-DFER, designed to effectively address the challenges inherent in weakly supervised dynamic facial expression recognition (DFER). By leveraging detailed, motion-based textual descriptions instead of conventional succinct labels, our method achieved a clearer semantic alignment between visual and textual modalities. This textual guidance, combined with a multi-grained temporal network, allowed our model to simultaneously capture fine-grained emotional nuances and broader temporal contexts, significantly enhancing the performance compared to existing MIL-based and CLIP-based approaches.

{
Our qualitative analyses, presented in Section 4.5, provide deeper insights into the model's behavior and performance characteristics. 
The confusion matrix analysis (Section 4.5.1) confirms that performance on certain classes, such as \textit{Disgust} and \textit{Fear}, is constrained by the severe data imbalance inherent in the benchmark datasets. 
Conversely, the influence visualizations (Section 4.5.2) qualitatively validate our core hypothesis that the model effectively learns to focus on the most emotionally salient frames to make its prediction. 
This demonstrates the success of our text-guided approach in navigating the ambiguity of in-the-wild expressions. 
These findings inform the following discussion of the framework's remaining challenges and future directions.
}

Although the TG-DFER framework exhibits significant achievements in DFER, several challenges remain.

\textbf{Data Imbalance:} The model struggles with expressions that are underrepresented in the training data, such as \textit{Disgust} and \textit{Fear}, leading to lower recognition accuracy for these expressions. This issue can be mitigated by implementing data augmentation and synthetic data generation to balance the dataset and provide a more comprehensive representation of all expressions.

\textbf{Overfitting to Specific Contexts:} Overfitting to specific datasets or contexts limits the model generalization. Variability in emotional expression across different individuals further complicates this issue. Utilizing diverse datasets that cover a wide range of scenarios and employing domain adaptation can enhance the robustness of the model across different situations.

\textbf{Occlusions and Variations:} Real-world conditions induce occlusions and variations in lighting and facial poses, which negatively affect the model performance. Implementing robust feature extraction methods or attention mechanisms that focus on the most informative parts of the frame can help to mitigate these issues.

\section{Conclusion}
In this paper, we proposed TG-DFER, an innovative weakly supervised method designed to enhance the accuracy and interpretability of dynamic facial expression recognition tasks. By integrating detailed, motion-based textual descriptions into a multi-grained temporal framework, our approach addresses the inherent limitations of coarse, video-level annotations. The introduction of textual guidance and enhanced fine-grained label substantially improved semantic alignment between visual and textual modalities.

The proposed method demonstrated significant performance improvements over existing MIL-based and CLIP-based methods in extensive experiments conducted on the DFEW and FERV39K datasets. Ablation studies further highlighted the effectiveness of combining local and global temporal information, as well as the critical role of enhanced fine-grained label in accurately identifying representative expression frames.

Future research could explore extending this approach to other emotion recognition tasks, incorporating additional modalities, and optimizing visual-textual alignment methods to further boost recognition performance in more challenging real-world scenarios.

\section{Acknowledgments}
This work was supported by the Institute for Information \& communications Technology Planning \& Evaluation(IITP) grant funded by the Korea government(MSIT) (No. RS-2019-II190079 (Artificial Intelligence Graduate School Program (Korea University)), No. IITP-2025-RS-2024-00436857 (Information Technology Research Center (ITRC))).

\end{document}